\documentclass[11pt,a4paper]{article}
\usepackage[hyperref]{eacl2021}
\usepackage{times}
\usepackage{latexsym}

\usepackage{graphicx}
\usepackage{enumitem}
\usepackage{arabtex}
\usepackage{utf8}
\setcode{utf8}

\usepackage{microtype}
\title{Arabic Offensive Language on Twitter:  Analysis and Experiments}
\author{
\AND
\begin{tabular}{ccc}
     Hamdy Mubarak $^1$&Ammar Rashed$^2$&Kareem Darwish$^1$
\end{tabular}
\AND
\begin{tabular}{cc}
     Younes Samih$^1$&Ahmed Abdelali$^1$  
\end{tabular}
\\
\\
\begin{tabular}{cc}
     $^1$Qatar Computing Research Institute, HBKU & $^2$\"{O}zye\u{g}in University\\
     \{hmubarak, kdarwish, ysamih, aabdelali\}@hbku.edu.qa& ammar.rasid@ozu.edu.tr

\end{tabular}
}

\date{}

\begin{document}
\aclfinalcopy
\maketitle
\begin{abstract}
Detecting offensive language on Twitter has many applications ranging from detecting/predicting bullying to measuring polarization. In this paper, we focus on  building a large Arabic offensive tweet dataset. We introduce a method for building a dataset that is not biased by topic, dialect, or target. We produce the largest Arabic dataset to date with special tags for vulgarity and hate speech.  We thoroughly analyze the dataset to determine which topics, dialects, and gender are most associated with offensive tweets and how Arabic speakers use offensive language. Lastly, we conduct many experiments to produce strong results (F1 = 83.2) on the dataset using SOTA techniques.
\end{list}
\end{abstract}

\section{Introduction}
\textbf{Disclaimer:} Due to the nature of the paper, some examples herein contain highly offensive language and hate speech. They don't reflect the views of the authors in any way. This work is an attempt to help fight such speech.\newline % None of these examples are directed at any groups or individuals. Conversely, we attempt to contribute to fight radical, racist and other toxic contents on the internet.\newline

Much recent interest has focused on the detection of offensive language and hate speech in online social media.  Offensiveness is often associated with undesirable behaviors such as trolling, cyberbullying, online extremism, political polarization, and propaganda. Thus, offensive language detection is instrumental for a variety of application such as: quantifying polarization \cite{barbera2015follow,conover2011political}, trolls and propaganda account detection \cite{darwish2017seminar}, hate crimes likelihood estimation \cite{waseem2016hateful}; and predicting conflicts \cite{chadefaux2014early}.
% Ammar adding this for consideration
% sorry ya Ammar, this is implied in our work already
% We hope this work potentially contributes to fighting toxicity on social media.
%
In this paper, we describe our methodology for building a large dataset of Arabic offensive tweets. Given that roughly 1-2\% of all Arabic tweets are offensive \cite{mubarak2019arabic}, targeted annotation is essential to efficiently build a large dataset.  Since our methodology does not use a seed list of offensive words, it is not biased by topic, target, or dialect. Using our methodology, we tagged a 10,000 Arabic tweet dataset for offensiveness, where offensive tweets account for roughly 19\% of the tweets.  Further, we labeled tweets as vulgar or hate speech. To date, this is the largest available dataset, which we plan to make  publicly available along with annotation guidelines. We use this dataset to characterize Arabic offensive language to ascertain the topics, dialects, and users' gender that %: what expressions are used in Arabic offensive language; who uses offensive language in terms of gender and dialect; and which topics 
are most associated with the use of offensive language.  Though we suspect that there are common features that span different languages and cultures, some characteristics of Arabic offensive language are language and culture specific.  Thus, we conduct a thorough analysis of how Arab users use offensive language.  Next, we use the dataset to train strong Arabic offensive language classifiers using state-of-the-art representations and classification techniques.  Specifically, we experiment with static and contextualized embeddings for representation along with a variety of classifiers such as Transformer-based and Support Vector Machine (SVM) classifiers. The contributions of this paper are as follows:  
\begin{itemize}
    \item We built the largest Arabic offensive language dataset to date that is also labeled for vulgar language and hate speech and is not biased by topic or dialect.  We describe the methodology for building it along with annotation guidelines.
    \item We performed thorough analysis to describe the peculiarities of Arabic offensive language.
    \item We experimented with SOTA classification techniques to provide strong results on detecting offensive language.
\end{itemize}
\section{Related Work}
Many recent papers have focused on the detection of offensive language, including hate speech \cite{agrawal2018deep,badjatiya2017deep,davidson2017automated,djuric2015hate,kwok2013locate,malmasi2017detecting,nobata2016abusive,yin2009detection}. Offensive language can be categorized as: \textit{vulgar}, which include explicit and rude sexual references, \textit{pornographic}, and \textit{hateful}, which includes offensive remarks concerning people’s race, religion, country, etc. \cite{jay2008pragmatics}.   Prior works have concentrated on building annotated corpora and training classification models.  Concerning corpora, \url{hatespeechdata.com} attempts to maintain an updated list of hate speech corpora for multiple languages including Arabic and English. Further, SemEval 2019 ran an evaluation task targeted at detecting offensive language, which focused exclusively on English \cite{zampieri2019semeval}. For SemEval 2020, they extended the task to include other languages including Arabic \cite{zampieri2020semeval}. As for classification models, most studies used supervised classification at either word level \cite{kwok2013locate}, character sequence level \cite{malmasi2017detecting}, and word embeddings \cite{djuric2015hate}. The studies used different classification techniques including Na{\"i}ve Bayes \cite{kwok2013locate}, Support Vector Machines (SVM) \cite{malmasi2017detecting}, and deep learning \cite{agrawal2018deep,badjatiya2017deep,nobata2016abusive} classification. The accuracy of the aforementioned system ranged between 76\% and 90\%.  Earlier work looked at the use of sentiment words as features as well as contextual features \cite{yin2009detection}.

The work on Arabic offensive language detection is relatively nascent \cite{abozinadah2017detecting,alakrot2018towards,albadi2018they,mubarak2017abusive,mubarak2019arabic}.  \newcite{mubarak2017abusive} suggested that certain users are more likely to use offensive languages than others, and they used this insight to build a list of offensive Arabic words and to construct a labeled set of 1,100 tweets.  \newcite{abozinadah2017detecting} used supervised classification based on a variety of features including user profile features, textual features, and network features.  They reported an accuracy of nearly 90\%. \newcite{alakrot2018towards} used supervised classification based on word n-grams to detect offensive language in YouTube comments.  They improved classification with stemming and achieved a precision of 88\%.  %Unfortunately, we don't have access to the datasets of Abozinadah et al. \cite{abozinadah2017detecting} and Alakrot et al. \cite{alakrot2018towards}.  
\newcite{albadi2018they} focused on detecting religious hate speech using a recurrent neural network. 

Arabic is a morphologically rich language with a standard variety called Modern Standard Arabic (MSA), which is typically used in formal communication, and many dialectal varieties that differ from MSA in lexical selection, morphology, phonology, and syntactic structures.  In MSA, words are typically derived from a set of thousands of roots by fitting a root into a stem template and the resulting stem may accept a variety of prefixes and suffixes.
%such as coordinating conjunctions and pronouns. 
Though word segmentation, which greatly improves word matching, is quite accurate for MSA \cite{abdelali2016farasa}, with accuracy approaching 99\%, dialectal segmentation is not sufficiently reliable, with accuracy ranging between 91-95\% for different dialects \cite{samih2017learning}. Since dialectal Arabic is ubiquitous in Arabic tweets and many tweets have creative spellings of words, recent work on Arabic offensive language detection used character-level models \cite{mubarak2019arabic}.  % We use character-level models in this paper. 

\section{Data Collection}
\subsection{Collecting Arabic Offensive Tweets}
Our target is to build a large Arabic offensive language dataset that is representative of its appearance on Twitter and is hopefully not biased to specific dialects, topics, or targets. % Building such a dataset is challenging.  
One of the main challenges is that offensive tweets constitute a very small portion of overall tweets. To quantify their proportion, we took 3 random samples of tweets from different days, with each sample composed of 1,000 tweets, and we found that only 1-2\% of them were offensive (including pornographic advertisements).  This percentage is consistent with previously reported percentages \cite{mubarak2017abusive}. Thus, annotating random tweets is grossly inefficient. One way to overcome this problem is to use a seed list of offensive words to filter tweets. However, doing so is problematic, as it would skew the dataset to particular types of offensive language or to specific dialects.  % There are multiple varieties of Arabic, including so-called Modern Standard Arabic (MSA), which is used in formal communications, and five major dialects, namely: Egyptian (EGY), Gulf (GLF), Levantine (LEV), Maghrebi (MGR), and Iraqi (IRQ), with more than 20 sub-dialects.
Offensiveness is often dialect and country specific. 

After inspecting many tweets, we observed that many offensive tweets have the vocative particle \<يا>  
(``yA'' -- meaning ``O'')\footnote{Arabic words are provided along with their Buckwalter transliteration and English translation.}, which is mainly used in directing the speech to a specific person or group.  
The ratio of offensive tweets increases to 5\% if a tweet contains one vocative particle and to 19\% if it has at least two vocative particles.  %we consider tweets having this particle. And if we consider tweets having at least two or of these particles and get unique tweets having word sequence in the middle, the ratio of having abusive tweets increases to 19\%. 
Users often repeat this particle for emphasis, as in:  \<يا أمي يا حنونة> 
(``yA Amy yA Hnwnp'' -- O my mother, O kind one), which is endearing and non-offensive, and 
\<يا كلب يا قذر> 
(``yA klb yA q*r'' -- ``O dog, O dirty one''), which is offensive. We decided to use this pattern to increase our chances of finding offensive tweets.  
% Some example usages of the vocative article include 
% \<يا ابني> 
% (``yA Abny'' -- ``O my son''), which is not offensive, and 
% \<يا غبي>  
% (``yA gby'' -- ``O dummy''), which is offensive. 
% \subsection{Pattern}
One of the main advantages of the pattern \<يا ... يا> 
(``yA ... yA'') is that it is not associated with any specific topic or genre, and it appears in all Arabic dialects. Though the use of offensive language does not necessitate the appearance of the vocative particle, the particle does not favor any specific offensive expressions and greatly improves our chances of finding offensive tweets.\\ %We used this pattern to increase existence of offensive tweets in the annotation dataset.
Using Twitter APIs, we collected 660k Arabic tweets having this pattern between April 15 -- May 6, 2019. To increase diversity, we sorted the word sequences between the vocative particles and took the most frequent 10,000 unique sequences.  For each word sequence, we took a random tweet containing that sequence. Then we annotated those tweets, ending up with 1,915 offensive tweets which represent roughly 19\% of all tweets. Each tweet was labeled as: offensive, which could additionally be labeled as vulgar and/or hate speech, or Clean. We describe in greater detail our annotation guidelines, which are compatible with the OffensEval2019 annotation guidelines ~\cite{zampieri2019semeval}. % We used this initial classification for better understanding and analysis of each types.  , then we merged vulgar and offensive tweets in one class (OFF) to be consistent with OffensEval2019 annotation guidelines ~\cite{zampieri2019semeval}. This can help in comparing usage of offensive language in Arabic with other languages (e.g. English). We also added two more labels to each tweet, %to be consistent with the same guidelines
 % namely: whether the tweet contains insult/threat to an individual or group (Targeted or Untargeted Insult), and whether the target of the offensive tweet is Individual or Group. 
 %These are Sub-tasks B and C in OffensEval2019 guidelines. 
 For example, if a tweet has insults or threats targeting a group based on their nationality, ethnicity, gender, political affiliation, religious belief, or other common characteristics, this is considered hate speech ~\cite{zampieri2019semeval}. It is worth mentioning that we also considered insulting groups based on their sport affiliation as a form of hate speech. Often, being a fan of a particular sporting club is considered a part of the personality that rarely changes over time (similar to religious and political affiliations). Many incidents of violence have occurred among fans of rival clubs. %\\
 %\\
 % \textcolor{red}{\textbf{Limitation:} 
 
 Although we used a generic pattern that is used across dialects and topics, such may not cover all the stylistic diversity of offensive expressions.  However, our approach considerably narrows the search space for offensive tweets, which constitute a small percentage of tweets in general, while being far more generic than using a seed list of offensive words, which may greatly skew the distribution of offensive tweets.  % overcomes problems associated with  e acknowledge that we didn't cover all linguistic styles of offensive language usage (e.g. offensive tweets without having vocative particles). 
 For future work, we plan to explore other methods for identifying offensive tweets with greater stylistic diversity. % data collection without exposing biases for future work.
 %}

%All experiments reported in this paper correspond to Sub-task A; identifying whether tweet is offensive (OFF) or not (NOT), and we leave detecting vulgar language and hate speech for future work.  
% Next, we present our annotation guidelines.
%\textcolor{red}{\HandRight It sounds too much overlap with OffensEval?? I don't think so. these are just the annotation guidelines}
%Then we annotated  tweets, prioritizing those with most frequent sequences first, until we had 2,000 offensive tweets and 8,000 non-offensive tweets, for a total of 10,000. 
%, ending up with unique 10k word sequences and unique 10k tweets ready for annotation and analysis.
% Usages of this pattern vary a lot; they can be prayer or supplication, ex: \<يا حي يا قيوم> (O God, O Lord), or direct speech to someone, ex: \<يا أمي يا حنونة>,(O mother, O kind), or attacking a person in offensive tweet, ex: \<يا كلب يا قذر> (O dog, O dirty) among others. 
% Frequency of yA Hy in our dataset = 13000 while frequency of ya klb = 400, (3\%), but if we take only one tweet from each word sequence, this ratio increased to 19\%, see following section.
%This pattern is also used in dialects for choices, ex: \<يا تسافر يا تقعد> (either you stay or you travel)

\subsection{Annotating Tweets}
We developed annotation guidelines jointly with an experienced annotator, who is a native Arabic speaker with good knowledge of various Arabic dialects, in accordance to the OffensEval2019 guidelines.    %
%The annotator carried out all annotation.  % , 
Tweets were given one or more of the following four labels: \textit{offensive}, \textit{vulgar}, \textit{hate speech}, or \textit{clean}.  Since the \textit{offensive} label covers both vulgar and hate speech and vulgarity and hate speech are not mutually exclusive, a tweet can be just offensive or offensive and vulgar and/or hate speech.  %The vulgar and offensive labels are not mutually exclusive. 
The annotation adhered to the following guidelines:
\paragraph{OFFENSIVE (OFF):}
Offensive tweets contain explicit or implicit insults or attacks against other people, or inappropriate language, such as: \\ % theses are some common examples:\\
\textbf{Direct threats or incitement}, ex: 
\<احرقوا مقرات المعارضة> 
(``AHrqwA mqrAt AlmEArDp'' -- ``burn opposition headquarters'') and 
\<اقتلوا هذا المنافق> 
(``h*A AlmnAfq yjb qtlh'' -- ``kill this hypocrite'').\\
\textbf{Insults and expressions of contempt}, which include: \textit{Animal analogies}, ex: 
\<يا كلب> 
(``yA klb'' -- ``O dog'') and 
\<كل تبن> 
(``kl tbn'' -- ``eat hay'').; \textit{Insult to family}, ex: 
\<يا روح أمك> 
(``yA rwH Amk'' -- ``O mother's soul''); 
%and \<أبو اللي جابك> 
%(``$>$bw Ally jAbk'' -- ``Is cursed the one who brought /beget you'')\\
\textit{Sexually-related insults}, ex: 
\<يا ديوث> 
(``yA dywv'' -- ``O cuckold''); 
%or \<الشواذ أمثالك> 
%(``Al\$wA* $>$mvAlk'' -- %``gays like you'')\\
\textit{Damnation}, ex: \<الله يلعنك>
(``Allh ylEnk'' -- ``may God curse you''); and 
\textit{Attacks on morals}, ex: 
\<يا كاذب> 
(``yA kA*b'' -- ``O liar'').
\paragraph{VULGAR (VLG):} Vulgar tweets are offensive tweets that contain profanity, such as mentions of private parts or sexual-related acts or references.
\paragraph{HATE SPEECH (HS):} Hate speech tweets are offensive tweets targeting group based on common characteristics such as: 
\textit{Race}, ex: \<يا زنجي> 
(``yA znjy'' -- ``O Negro''); \textit{Ethnicity}, ex. \<الفرس الأنجاس> 
(``Alfrs AlAnjAs'' -- ``Impure Persians''); 
\textit{Group or party}, ex: 
\<أبوك شيوعي> (``Abwk \$ywEy'' -- ``your father is a communist''); and \textit{Religion}, ex: 
\<دينك القذر> 
(``dynk Alq*r'' -- ``your filthy religion'').
\paragraph{CLEAN (CLN):}  Clean tweets do not contain vulgar or offensive language.  We noticed that some tweets have some offensive words, but the whole tweet should not be considered as offensive due to the intention of users. This suggests that normal string match without considering contexts may fail in some cases. Examples of such ambiguous cases include:  % addition to the following cases:\\
\textit{Humor}, ex: 
\<يا عدوة الفرحة ههه> 
(``yA Edwp AlfrHp hhh'' -- ``O enemy of happiness hahaha''); 
\textit{Advice}, ex: \<لا تقل لصاحبك يا خنزير> 
(``lA tql lSAHbk yA xnzyr'' -- ``don't say to your friend: You are a pig''); 
\textit{Condition}, ex: \<إذا عارضتهم يقولون يا عميل> 
(``A*A EArDthm yqwlwn yA Emyl'' -- ``if you disagree with them, they call you a spy''); 
\textit{Condemnation}, ex: \<لماذا نسب بقول: يا بقرة؟> 
(``lmA*A nsb bqwl: yA bqrp?'' -- ``Why do we insult others by saying: O cow?''); 
\textit{Self offense}, ex: \<تعبت من لساني القذر> 
(``tEbt mn lsAny Alq*r'' -- ``I am tired of my dirty tongue''); 
\textit{Non-human target}, ex: \<يا بنت المجنونة يا كورة> 
(``yA bnt Almjnwnp yA kwrp'' -- ``O daughter of the crazy one O football''); and 
\textit{Quotation from a movies or a story}, ex: \<تاني يا زكي! تاني يا فاشل> 
(``tAny yA zky! tAny yA fA\$l'' -- ``again smarty! again O loser''). For ambiguous expressions, the annotator searched Twitter to observe real sample usages.

Table \ref{tab:offensiveLanguageDistribution} shows the distribution of the annotated tweets.  There are 1,915 offensive tweets, including 225 vulgar tweets and 506 hate speech tweets, and 8,085 clean tweets. To validate annotation quality, we asked three additional annotators to annotate two tweet sample sets.  The first was a random sample of 100 tweets containing 50 offensive and 50 non-offensive tweets. The Inter-Annotator Agreement (IIA) between the annotators using Fleiss’s Kappa coefficient~\cite{fleiss1971measuring} was 0.92. % \textcolor{red}{
The second was general random samples containing 100 tweets each from the dataset, and the IIA with the dataset was: 0.97, 0.96, and 0.97. This high level of agreement gives more confidence in the  quality of the annotation. % and agreement. % }.\\
\textcolor{black}{Data can be downloaded from:\\ \url{https://alt.qcri.org/resources/OSACT2020-sharedTask-CodaLab-Train-Dev-Test.zip}}

\begin{table}[pt]
    \centering
    \begin{tabular}{l|r|r} % |c}
         & Tweets & Words \\ \hline % & Words/Tweet\\ \hline
        Offensive & 1,915 & 38k \\ % & 23\\
        -- Vulgar & 225 & 4k \\ %  & 18\\
        -- Hate speech & 506 & 13k \\ %  & 26 \\  %% since we have space, we should use the full name.  it makes the paper much more readable.
        Clean & 8,085 & 151k \\ \hline %  19\\ \hline
        Total & 10,000 & 193k % & 19
    \end{tabular}
    \caption{Distribution of offensive and clean tweets.} % Hate Speech tweets are parts of offensive and vulgar tweets.}
    \label{tab:offensiveLanguageDistribution}
    % \squeezeup
\end{table}

\subsection{Statistics and User Demographics}
\begin{figure}[h]
  \centering
\includegraphics[width=.9\linewidth]{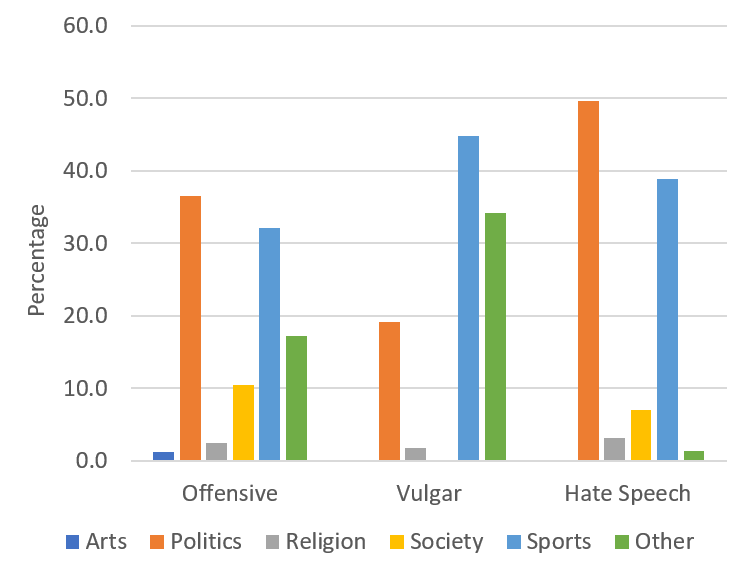}
    \caption{Topic distribution for offensive language and its sub-categories}
    \label{fig:topicDistribution}
\end{figure}

\begin{figure}[h]
  \centering
    \includegraphics[width=.9\linewidth]{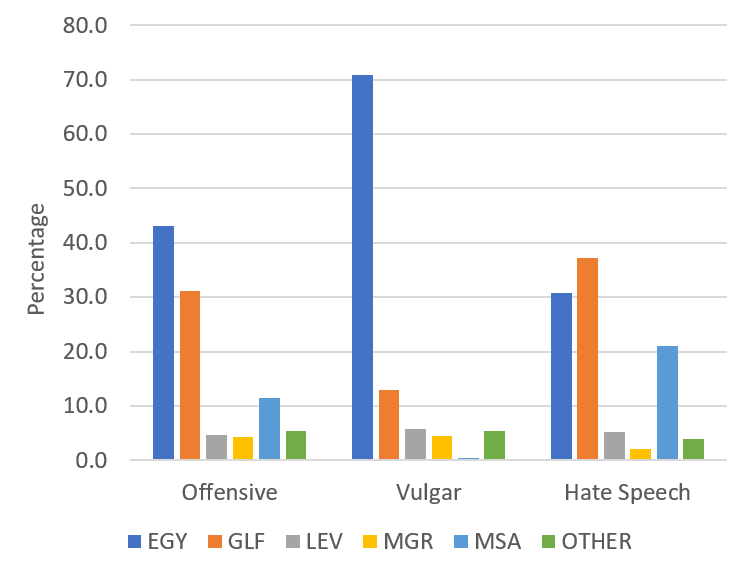}
    \caption{Dialect distribution for offensive language and its sub-categories}
    \label{fig:dialectDistribution}
\end{figure}

\begin{figure}[ht]
    \centering
    \includegraphics[width=0.85\linewidth]{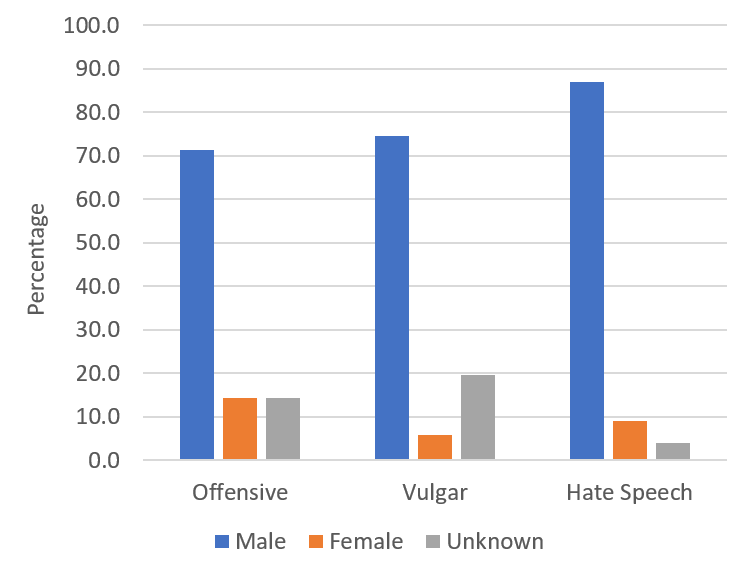}
    \caption{Gender distribution for offensive language and its sub-categories}
    \label{fig:genderDistribution}
    % \squeezeup
\end{figure}

\begin{figure}[!tbp]
  \centering
\includegraphics[width=\linewidth]{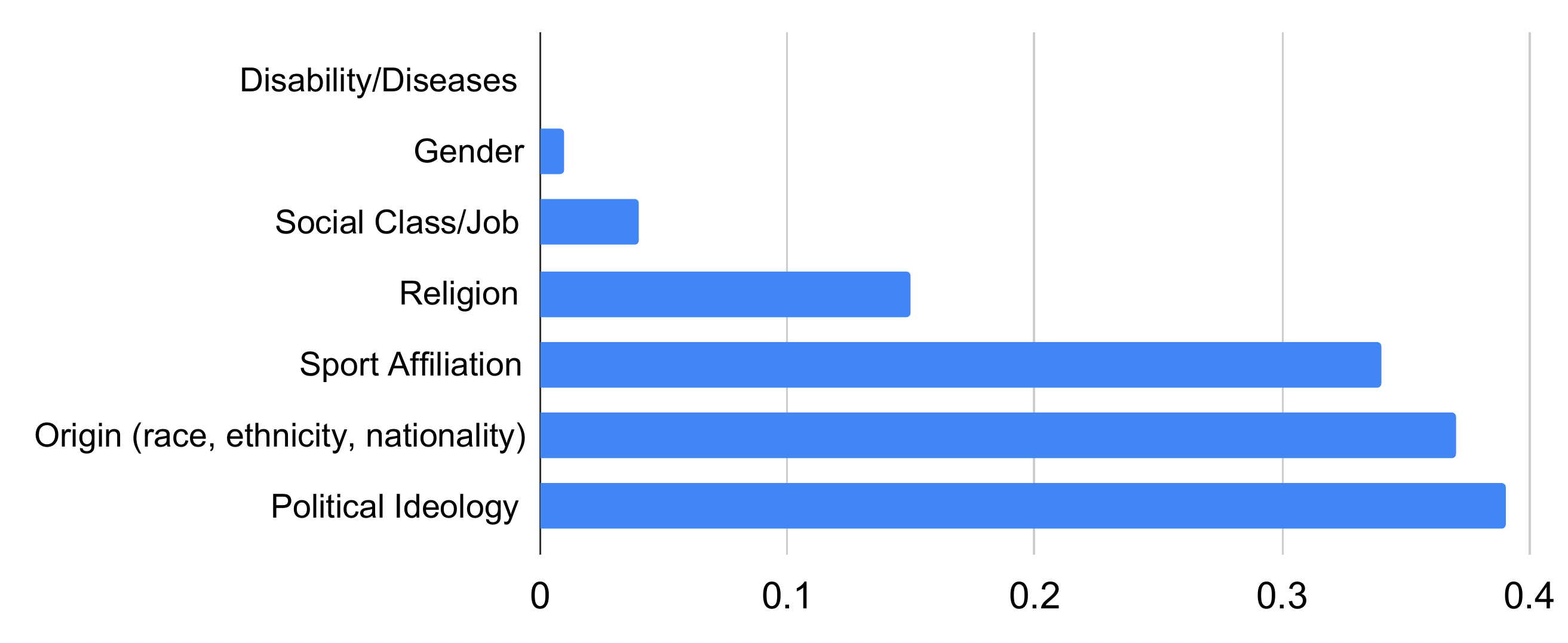}
%}
\caption{Distribution of Hate Speech Types. Note: A tweet may have more than one type.}
\label{fig:HateSpeechTypes}
\end{figure}

Given the annotated tweets, we wanted to ascertain the distribution of: types of offensive language, genres or topics where it is used, the dialects used, and the gender of users using such language.  Accordingly, the annotator manually examined and tagged all the offensive tweets.\newline
\textbf{Topic:} Figure \ref{fig:topicDistribution} shows the distribution of topics associated with offensive tweets.  As the figure shows, sports and politics are most dominant for offensive language including vulgar and hate speech.  \newline
\textbf{Dialect:} We looked at MSA and four major dialects, namely Egyptian (EGY), Leventine (LEV), Maghrebi (MGR), and Gulf (GLF).  Figure \ref{fig:dialectDistribution} shows that 71\% of vulgar tweets were written in EGY followed by GLF, which accounted for 13\% of vulgar tweets. % we found that majority of vulgar tweets are written in Egyptian (EGY) dialect (71\%) while the Gulf dialect (GLF) was used in 13\% of the cases. 
MSA was not used in any vulgar tweets.  As for offensive tweets in general, EGY and GLF were used in 36\% and 35\% of the offensive tweets respectively.  Unlike the case of vulgar language, 15\% of the offensive tweets were written in MSA. For hate speech, GLF and EGY were again dominant and MSA constituted 21\%  of the tweets. % while other dialects including MSA were used in 28\% of the communications.
This is consistent with findings for other languages, e.g. English and Italian, where vulgarity was more frequently associated with colloquial language ~\cite{mattiello2005pervasiveness,maisto2017mining}. \newline
\textbf{Gender:} Figure \ref{fig:genderDistribution} shows that the vast majority of offensive tweets, including vulgar and hate speech, were authored by males.  Female Twitter users accounted for 14\% of offensive tweets in general and 6\% and 9\% of vulgar and hate speech respectively.\newline
Figure \ref{fig:HateSpeechTypes} shows a detailed categorization of hate speech types, where the top three include insulting groups based on their political ideology, origin, and sport affiliation. Religious hate speech appeared in only 15\% of all hate speech tweets.

%Figure XX shows their distribution for vulgar and offensive language. Sports domain was the top domain that has vulgar language (44\%) while Politics domain appears in 14\% of the vulgar tweets. For Offensive language, Politics and Sports domains have 41\% and 29\% in order, and other domains represent 30\%.\\
% Gender classification: we found that 75\% of the vulgar tweets are written by Male users, while Female users wrote 6\%, and in 20\% of the cases, the gender was not defined. In offensive language, tweets written by Male and Female speakers represent 69\% and 10\% in order.\\

% \begin{figure}[ht]
%     \centering
    
%     \squeezeup
% \end{figure}
% \begin{figure}[ht]
%     \centering

%     \squeezeup
% \end{figure}

\begin{figure}[!tbp]
  \centering
\includegraphics[width=.7\linewidth, height=.7\linewidth]{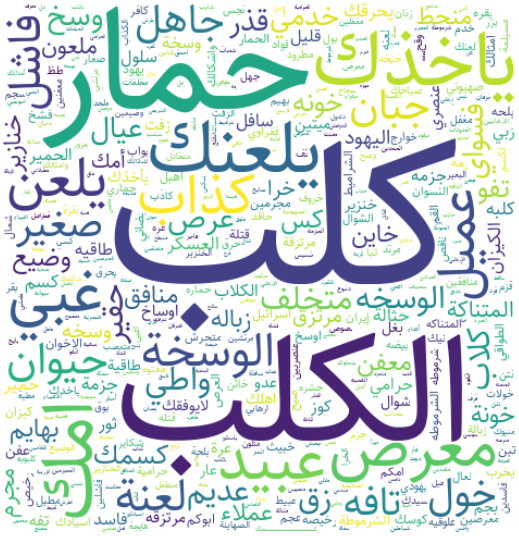}
%}
\caption{Tag cloud for words with top valence score among offensive class, e.g. name calling (animals), curses, insults, etc.}
\label{fig:topOffensiveWordCloud}
\end{figure}

% \begin{figure}[ht]
% \begin{center}
% %\frame{
% \includegraphics[width=0.6\linewidth]{figures/HateSpeechTypes.pdf}
% %}
% \caption{Distribution of Hate Speech Types. Note: A tweet may have more than one type.}
% \label{fig:HateSpeechTypes}
% \end{center}
% %\squeezeup
% \end{figure}
% \subsection{Analysis of Offensive Language Usage}
% We analyzed vulgar and offensive tweets. 
% From Arabic Wikipedia:
% \<لفظ ناب> Profanity:
% \<العربية: الكفر، الجنس، المثلية، التلفظ على أفراد العائلة كالأم والأخت، التشبيه بالحيوانات، الكره العنصري، السباب بالدين.>
% In other cultures, 
% \<الأعضاء التناسلية، الأمراض، اللعن، الكفر، إهانة الذكاء، الأمراض العقلية، الإفرازات الجسدية>
% We show that all the last cases are used on Twitter. \newline
% Sometimes tweets having offensive words but are not considered as offensive as shown in \label{fig:confusingCases}.
% \begin{figure}[pt]
% \begin{center}
% \frame{\includegraphics[width=0.9\linewidth]{confusingCases}}
% \caption{Confusing Cases}
% \end{center}
% \label{fig:confusingCases}
% \end{figure}
% \section{Analysis of Offensive Language Usage}
%(we apologize for the offensive nature of the examples): \\ % According to our dataset, below are our observations which can be added as special cases to the general annotation guidelines described before (in addition to MSA, people use their equivalent dialect variations) and these are just examples because Arabic is a very rich language and opinions can be expressed in many different ways .\\
Next, we analyzed all tweets labeled as offensive to better understand how Arabic speakers use offensive language. Here is a breakdown of usage:\newline
\textbf{Direct name calling:} 
The most frequent attack is to call a person an animal name, and the most used \textit{animals} were 
\<كلب> (``klb'' -- ``dog''), 
\<حمار> (``HmAr'' -- ``donkey''), and \<بهيم> (``bhym'' -- ``beast''). 
The second most common was insulting mental abilities using words such as \<غبي> (``gby'' -- ``stupid'') and 
\<عبيط> (``EbyT'' --``idiot''). 
Culturally, not all animal names are used as insults.  For example, animals such as 
\<أسد> (``Asd'' -- ``lion''),
\<صقر> (``Sqr'' -- ``falcon''), and 
\<غزال> (``gzAl'' -- ``gazelle'') are typically used for praise. %Generalization and copying from other cultures are not always valid. 
%Also, there are some differences between different regions in the Arab world in using offensive words.  For example, \<حصان> (``HSAn'' -- ``horse'') is considered offensive in some countries and as praise in others.  
% , for example some words are considered as offensive in some countries but normal or less offensive in the other countries, or even in different cities inside the same country.\\
For other insults, people use: some \textit{bird} names such as
\<دجاجة> 
(``djAjp'' -- ``chicken''),
\<بومة> 
(``bwmp'' -- ``owl''), and
\<غراب> 
(``grAb'' -- ``crow'');  \textit{insects} such as 
\<ذبابة> 
(``*bAbp'' -- ``fly''), 
\<صرصور>  
(``SrSwr'' -- ``cockroach''), and 
\<حشرة> 
(``H\$rp'' -- ``insect''); \textit{microorganisms} such as 
\<جرثومة> 
(``jrvwmp'' -- ``microbe'') and 
\<طحالب> (``THAlb'' -- ``algae''); \textit{inanimate objects} such as 
\<جزمة> (``jzmp'' -- ``shoes'') and 
\<سطل> (``sTl'' -- ``bucket'') among other usages.\\
%and some plants such as 
%\<بطيخ> (``bTyx'' -- ``watermelon'') and \<بصل> (``bSl'' -- ``onion'').
%\\
%and even some plants \<بصل، بطيخ> (onion, watermelon), or
% Another common offense is to describe a person as inanimate, ex: \<جزمة، دلو> (shes, bucket)\\
\textbf{Simile and metaphor:} Users use simile and metaphor were they would compare a person to: an \textit{animal} as in 
\<زي الثور> (``zy Alvwr'' -- ``like a bull''), 
\<سمعني نهيقك> (``smEny nhyqk'' -- ``let me hear your braying''), and 
\<هز ديلك> (``hz dylk'' -- ``wag your tail''); 
a person with \textit{mental or physical disability} such as 
\<منغولي> (``mngwly'' -- ``Mongolian (Down syndrome)''), 
\<معوق> (``mEwq'' -- ``disabled''), and
\<قزم> (``qzm'' -- ``dwarf'');  
and to the \textit{opposite gender} such as 
\<جيش نوال> (``jy\$ nwAl'' -- ``Nawal's army (Nawal is female name)'') and 
\<نادي زيزي> (``nAdy zyzy'' -- ``Zizi's club (Zizi is a female nickname)'').\\
\textbf{Indirect speech:} This includes: \textit{sarcasm} such as 
\<أذكى إخواتك> (``A*kY AxwAtk'' -- ``smartest one of your siblings'') and 
\<فيلسوف الحمير> (``fylswf AlHmyr'' -- ``the donkeys' philosopher''); 
\textit{questions} such as 
\<ايه كل الغباء ده> (``Ayh kl AlgbA dh'' -- ``what is all this stupidity''); 
and \textit{indirect speech} such as 
\<النقاش مع البهايم غير مثمر> (``AlnqA\$ mE AlbhAym gyr mvmr'' -- ``no use arguing with cattle''). \\
\textbf{Wishing Evil:} This entails wishing death or major harm to befall someone such as 
\<ربنا ياخدك> (``rbnA yAxdk'' -- ``May God take (kill) you''),
\<الله يلعنك> (``Allh ylEnk'' -- ``may God curse you''), and 
\<روح في داهية> (``rwH fy dAhyp'' -- equivalent to ``go to hell'').\\ 
\textbf{Name alteration:} One common way to insult others is to change a letter or two in their names to produce new offensive words that rhyme with the original names.  Some such examples include changing 
\<الجزيرة> (``Aljzyrp'' -- ``Aljazeera (channel)'') to 
\<الخنزيرة> (``Alxnzyrp'' -- ``the pig'') and
\<خلفان> (``xlfAn'' -- ``Khalfan (person name)'') to 
\<خرفان> (``xrfAn'' -- ``crazed'').\newline
\textbf{Societal stratification:} Some insults are associated with: certain \textit{jobs} such as 
\<بواب> (``bwAb'' -- ``doorman'') or 
\<خادم> (``xAdm'' -- ``servant'');
and specific \textit{societal components} such 
\<بدوي> (``bdwy'' -- ``bedouin'') and 
\<فلاح> (``flAH'' -- ``farmer''). \\
\textbf{Immoral behavior:} These insults are associated with negative moral traits or behaviors such as 
\<حقير> (``Hqyr'' -- ``vile''), 
\<خاين> (``xAyn'' -- ``traitor''), and 
\<منافق> (``mnAfq'' -- ``hypocrite''). \\
\textbf{Sexually related:} They include expressions such as 
\<خول> (``xwl'' -- ``gay''),
\<وسخة> (``wsxp'' -- ``prostitute''), and 
\<عرص> (``ErS'' -- ``pimp'').

Figure \ref{fig:topOffensiveWordCloud} shows the top words with the highest valance scores for individual words in the offensive tweets. Larger fonts are used to highlight words with highest scores and align as well with the categories mentioned in the breakdown for the offensive languages.  We slightly modified the valence score described by \cite{conover2011political} to magnify its value by multiplying valence with frequency of occurrence. 

\section{Experiments}
We conducted an extensive battery of experiments on the dataset to establish strong Arabic offensive language classification results.  Though offensive tweets have finer-grained labels where offensive tweet could also be vulgar and/or hate speech, we conducted coarser-grained classification to determine if a tweet was offensive or not. For classification, we experimented with several tweet representation and classification models.  For tweet representations, we used: the count of positive and negative terms, based on a polarity lexicon; static embeddings, namely fastText and Skip-Gram; and deep contextual embeddings, namely BERT\textsubscript{base-multilingual} and AraBERT \cite{antoun2020arabert}.

% ========AMMAR START========\\
\subsection{Data Pre-processing}
We performed several text pre-processing steps.  First, we tokenized the text using the Farasa Arabic NLP toolkit \cite{abdelali2016farasa}. Second, we removed URLs, numbers, and all tweet specific tokens, namely mentions, retweets, and hashtags as they are not part of the language semantic structure, and therefore, not usable in pre-trained embeddings. Third, we performed basic Arabic letter normalization, namely variants of the letter \textit{alef} to \textit{bare alef}, \textit{ta marbouta} to \textit{ha}, and \textit{alef maqsoura} to \textit{ya}. We also separated words that are commonly incorrectly attached such as \<ياكلب> 
(``yAklb'' -- ``O dog''), is split to 
\<يا كلب> 
(``yA klb'').
Lastly, we normalized letter repetitions to allow for a maximum of 2 repeated letters.  For example, the token \<ههههه> 
(``hhhhh'' -- ``hahahahaha'') is normalized to \<هه> 
(``hh''). We also removed Arabic diacritics and word elongations (kashida). 
% to express laughter.
%provide strong signal to classification models as we show in the Features subsection\ref{sub:Features}. 
% We removed network-based features, such as .
% We also evaluated fastText text classification with and without network-based features, and there was almost no difference in performance.
\subsection{Representations}
\paragraph{Lexical Features}\label{sub:Features}
Since offensive words typically have a negative polarity, we wanted to test the effectiveness of using a polarity lexicon in detecting offensive tweets.
For the lexicon, we used NileULex \cite{nileulex2016}, which is an Arabic polarity lexicon containing 3,279 MSA and 2,674 Egyptian terms, out of which 4,256 are negative and 1,697 are positive. We used the counts of terms with positive polarity and terms with negative polarity in tweets as features.

% For the first representation, we used the counts of postiterms in a tweet that, which were offensive words appearing the NileULex lexicon \cite{nileulex2016}.  NileULex contains a large inventory of MSA offensive words. Negation was handled to determine the polarity of the keyword in the tweet.  We used the number of positive and negative keywords as lexicon-based features.

%Although these features in table \ref{tab:features} perform reasonably well on their own, combining them does not add much to the performance of using embeddings features only.

\paragraph{Static Embeddings}
We experimented with various static embeddings that were pre-trained on different corpora with different vector dimensionality. We compared pre-trained embeddings to embeddings that were trained on our dataset. 
For pre-trained embeddings, we used: \textbf{fastText} Egyptian Arabic pre-trained embeddings \cite{fasttext2017} with vector dimensionality of 300; 
\textbf{AraVec} skip-gram embeddings \cite{AraVec2017}, trained on 66.9M Arabic tweets with 100-dimensional vectors; and 
\textbf{Mazajak} skip-gram embeddings \cite{Mazajak2019}, trained on 250M Arabic tweets with 300-dimensional vectors. Sentence embeddings were calculated by taking the mean of the embeddings of their tokens. The importance of testing a character level n-gram model like fastText lies in the agglutinative nature of the Arabic language. We trained a new fastText text classification model \cite{fasttextClassification2017} on our dataset with vectors of 40 dimensions, 0.5 learning rate, 2$-$10 character n-grams as features, for 30 epochs. These hyper-parameters were tuned using a 5-fold cross-validated grid-search.
% Using a classifier model like SVM using sentence as embeddings, we found that using the pre-trained Egyptian Arabic fastText embeddings provided by Facebook perform almost the same as new fastText embeddings trained on our dataset. AraVec and Mazajak are Skip-Gram models trained on a 66.9M and 250M Arabic tweets with vector dimensionality of 100 and 300 respectively. 

\paragraph{Deep Contextualized Embeddings}
We also experimented with pre-trained contextualized embeddings with fine-tuning for down-stream tasks. Recently, deep contextualized language models such as BERT (Bidirectional Encoder Representations from Transformers)~\cite{devlin-2019-bert}, UMLFIT~\cite{howard-ruder-2018}, and OpenAI GPT ~\cite{radford2018}, have achieved ground-breaking results in many NLP classification and language understanding tasks. In this paper, we fine-tuned BERT\textsubscript{base-multilingual} (or simply BERT) and AraBERT embeddings to classify Arabic offensive language on Twitter as it eliminates the need for feature engineering. % heavily engineered task-specific architectures. 
% Ammar adding this, tagged for double checking :D
Although Robustly Optimized BERT (RoBERTa) embeddings perform better than (BERT\textsubscript{large}) on GLUE \cite{wang2018glue}, RACE \cite{lai2017race}, and SQuAD \cite{rajpurkar2016squad} tasks, pre-trained multilingual RoBERTa models are not available.
% end 104 languages, 12-layer, 768-hidden, 12-heads, 110M parameters
BERT is pre-trained on Wikipedia text from 104 languages, and AraBERT is trained on a large Arabic news corpus containing 8.5M articles composed of roughly 2.5B tokens. Both use identical architectures and come with hundreds of millions of parameters. Both contain an encoder with $12$ Transformer blocks, hidden size of $768$, and $12$ self-attention heads. These embedding use BP sub-word segments. Following \newcite{devlin-2019-bert}, the classification consists of introducing a dense layer over the final hidden state $h$ corresponding to first token of the sequence, [CLS], adding a softmax activation on the top of BERT to predict the probability  of the $l$ label: %\\ \squeezeup $
$p(l|h) = softmax(Wh),$ %$ \squeezeup \\ 
where $W$ is the task-specific weight matrix. During fine-tuning, all BERT/AraBERT parameters together with $W$ are optimized end-to-end to maximize the log-probability of the correct labels.

% \begin{figure}[h]
%     \centering
%     \includegraphics[scale=.36]{figures/features.png}
%     \caption{Performance of combining features. Skip-Gram refers to Mazajak embeddings. Arrows point to the highest scores.}
%     \label{fig:my_label}
% \end{figure}

\subsection{Classification Models}
We explored different classifiers.  When using lexical features and %context-independent
pre-trained static embeddings, we primarily used an SVM classifier with a radial basis function kernel. Only when using the Mazajak embeddings, we experimented with other classifiers such as AdaBoost and Logistic regression. %, table \ref{tab:classification_models}. 
The SVM classifier performed the best on static embeddings, and we picked the Mazajak embeddings because they yielded the best results among all static embeddings. We used the Scikit Learn implementations of all the classifiers such as libsvm for the SVM classifier.  We also experimented with fastText, which trained embeddings on our data. When using contextualized embeddings, we fine-tuned BERT and AraBERT by adding a fully-connected dense layer followed by a softmax classifier, minimizing the binary cross-entropy loss function for the training data. For all experiments, we used the PyTorch\footnote{\url{https://pytorch.org/}} implementation by HuggingFace\footnote{\url{https://github.com/huggingface/transformers}} as it provides pre-trained weights and vocabularies. %, and build on top of that.
% wanted to compare different feature combinations with different classification models to produce a more reliable baseline. C-Support Vector classifier with $C = 1$, and an Radial Basis Function (RBF) kernel performed the best among all the classifier models we tested. We wanted to compare different embeddings and features so we injected them into our classifier models and the best performing static feature set was Mazajak embeddings. Limited by space, we show only the performance of the classification models we trained on the best performing features in table \ref{tab:classification_models}.

\subsection{Evaluation} % or data balancing?
%balancing
For all of our experiments, we used 5-fold cross validation with identical folds for all experiments.  Table \ref{tab:embedding_classification} reports on the results of using lexical features, static pre-trained embeddings with an SVM classifier, embeddings trained on our data with fastText classifier, and BERT and AraBERT over a dense layer with softmax activation. As the results show, using fine-tuned AraBERT yielded the best results overall, followed closely by Mazajak/SVM, with large improvements in precision over using BERT. The success of AraBERT was surprising given that it was not trained on social media text.  Perhaps, pre-training a Transformer model on social media text may improve results further.  We suspect that the Mazajak/SVM combination performed better than BERT due to the fact that the Mazajak embeddings, though static, were trained on in-domain data, as opposed to BERT. % Perhaps if BERT embeddings were trained on tweets, they might have outperformed all other setups. 
For completeness, we compared 7 other classifiers with SVM using Mazajak embeddings. As results in Table \ref{tab:classification_models} show, using SVM yielded the best results.
% COMPARING BERT WITH pre-trained fastText
% end
% Nonetheless, the BERT results were better than all of them.

% The ratio of our data is $1:4$ offensive to not offensive tweets. To account for the imbalance in the data we report all our metrics on offensive class only. All our results are obtained with 5-fold cross validation to account for the small amount of data. We split our data into 5 folds (each has 2000 tweets) with the same distribution of classes.
% balancing
% In order to test the effect of bias in the data, we added three duplicates of all the under-represented offensive tweets to the data and retrained our models on the balanced version. The balancing effect on performance is shown in table \ref{tab:embedding_classification}.
% and plotted in figure  \ref{fig:oversampling_performance}. 

% \begin{table}[]
%     \centering
%     \begin{tabular}{l|l|l}
%          Group & Feature & F1\\
%          \hline
%          \multirow{4}{*}{Network} &  \# of hastags.&\multirow{4}{*}{45.03}\\
%          & Is a retweet.\\
%          & \# of URLs.\\
%          &\# of mentions.\\
%          \hline
%         %  \multirow{3}{*}{Linguistic}&  \# of tokens.&\multirow{3}{*}{44.47}\\
%         %  &  \# of emojis.\\
%         %  &  \# of callings.\\
%         %  \hline
%          \multirow{2}{*}{Lexicon}& \# of $+$ keywords.&\multirow{3}{*}{\textbf{68.54}}\\
%          & \# of $-$ keywords.\\
         
%     \end{tabular}
%     \caption{List of features used besides embeddings. F1 is measured on offensive class only.}
%     \label{tab:features}
% \end{table}

\begin{table}[h]
    \centering
%     \begin{tabular}{l||r|r|r|r}
% \hline
% Model &  Accuracy &  Precision &  Recall &  F1 \\
% \hline
% fastText Classifier  &     90.44 &      \textbf{77.81} &   73.06 &       75.26 \\
% SVM (fastText)       &     86.37 &      64.13 &   71.79 &       67.72 \\
% SVM (AraVec SG 100)  &     89.66 &      70.23 &   83.56 &       76.31 \\
% SVM (Mazajak SG 250) &     \textbf{91.68} &      76.67 &   \textbf{83.91} &       \textbf{80.11} \\
\begin{tabular}{l||r|r|r}
\hline
Model/classifier  &  Prec. &  Recall &  F1 \\
\hline
% \parbox[t]{2mm}{\multirow{5}{*}{\rotatebox[origin=c]{90}{Original}}}&
\multicolumn{4}{c}{Lexical Features} \\ \hline
SVM & 68.5 &  35.3 &  46.6\\ \hline
\multicolumn{4}{c}{Pre-trained static embeddings} \\ \hline
fastText/SVM & 76.7 &  43.5 &     55.5 \\
AraVec/SVM        & 85.5 &  69.2 &     76.4 \\
Mazajak/SVM    & \textbf{88.6} &  72.4 &     79.7 \\ \hline
\multicolumn{4}{c}{Embeddings trained on our data} \\ \hline
fastText/fastText      & 82.1 &  68.1 &     74.4 \\ \hline
% \multicolumn{4}{c}{Pre-trained Sentence Encoders} \\ \hline
% MUSE\textsubscript{V3}/SVM & 81.0 & 30.6 & 44.2 \\ \hline
\multicolumn{4}{c}{Contextualized embeddings} \\ \hline
% BERT\textsubscript{base-multilingual-uncased}     & \textbf{90.6}	&	\textbf{90.7}	&	\textbf{90.6} \\
BERT\textsubscript{base-multilingual} & 78.3	&	74.0	&	76.0 \\
AraBERT & 84.6	&	\textbf{82.4}	&	\textbf{83.2} \\
% AraBERT\textsubscript{base} & 84.6	&	\textbf{82.4}	&	\textbf{83.2} \\
\hline

% \parbox[t]{2mm}{\multirow{4}{*}{\rotatebox[origin=c]{90}{Balanced}}}&SVM(lexicon)&50.6&66.6&57.5\\
% &SVM (fastText) &   56.3 &  73.5 &     63.7 \\
% &fastText       &   \textbf{77.8} &  73.1 &     75.3 \\
% &AraVec         &   70.2 &  83.6 &     76.3 \\
% &Mazajak        &   76.7 &  \textbf{83.9} &     \textbf{80.1} \\
% \hline

\end{tabular}

    \caption{Classification performance with different features and models. %language models. SVM(fastText) refer to using SVM on the off-the-shelf pre-trained Egyptian Arabic fastText embeddings. fastText refer to using fastText text classification API, which uses a softmax layer. AraVec and Mazajak refer to using SVM on each of their embeddings.
    }
    \label{tab:embedding_classification}
% \squeezeupless
\end{table}

\begin{table}[h]
    \centering
    \begin{tabular}{l|r|r|r}
    \hline
         Model & Prec. & Recall & F1 \\
         \hline
         Decision Tree       &      51.2 &   53.8 &  52.4 \\
Random Forest       &      82.4 &   42.4 &  56.0 \\
Gaussian NB         &      44.9 &   \textbf{86.0} &  59.0 \\
Perceptron          &      75.6 &   67.7 &  66.8 \\
AdaBoost            &      74.3 &   67.0 &  70.4 \\
Gradient Boosting   &      84.2 &   63.0 &  72.1 \\
Logistic Regression &      84.7 &   69.5 &  76.3 \\
SVM                 &      \textbf{88.6} &   72.4 &  \textbf{79.7} \\
\hline
    \end{tabular}
    \caption{Performance of different classification models on Mazajak embeddings.}
    \label{tab:classification_models}
% \squeezeupless
\end{table}

% \begin{figure}[h]
%     \centering
%     \includegraphics[scale=.27]{figures/wlexicon.png}
%     \caption{Performance of the five baseline models. Thinner red bars represent the performance on the version of the data balanced by oversampling offensive instances.}
%     \label{fig:oversampling_performance}
% \end{figure}

% ========AMMAR END==========\\
% We split data into 5 folds (each has 2000 tweets) and we used fastText trained on 4 parts and tested on the remaining part. We used character windows ranging between 2 and 6 characters, and vector dimension of  100 and learning rate of 0.05. Training was repeated for 50 epochs.
% Table \ref{tab:5foldsAccuracy} shows results of the 5 folds.
% \begin{table}[pt]
%     \centering
%     \begin{tabular}{r|c}
%          & Accuracy \\ \hline
%         Fold1 & 90.3\% \\
%         Fold2 & 90.1\%\\
%         Fold3 & 90.1\%\\
%         Fold4 & 90.7\% \\
%         Fold5 & 89.3\%
%     \end{tabular}
%     \caption{5-fold fastText Accuracy}
%     \label{tab:5foldsAccuracy}
% \end{table}
% - fastText for EGY dataset (VL and OF)
% - fastText for the original tweet
% - for normalized text
% - all mentions are converted into @USER
% - all URLs are converted into URL\newline 
% \subsection{Hatebase.org} Comparison  with Hatebase.org
% In our work, we use fastText, which is an effective deep learning text classifier. The focus in this work on vulgar and pornographic speech in Arabic social media. 
\subsection{Error Analysis}
We inspected the tweets of one fold that were misclassified by the Mazajak/SVM model (36 false positives/121 false negatives) to determine the most common errors. They were as follows:
%with 90\% or more confidence
%to investigate the source of most prominent error signals. There were 36 false positives and 121 false negatives.
\paragraph{Four false positive types:} 
% We randomly selected 100 errors, we ascertain the most common error types:
% had four main types: % of false-positive:
\begin{itemize}[leftmargin=*]
    \item \textbf{Gloating:} ex. \<يا هبيده>
    (``yA hbydp'' - ``O you delusional'') referring to fans of rival sports team for thinking they could win.
    \item \textbf{Quoting:} ex. \<لما حد يسب ويقول يا كلب> 
    (``lmA Hd ysb wyqwl yA klb'' -- ``when someone swears and says: O dog''). %, where the tweet quotes a statement.
    \item \textbf{Idioms:} ex. \<يا فاطر رمضان يا خاسر دينك> 
    (``yA fATr rmDAn yA xAsr dynk'' -- ``o you who does not fast Ramadan, you have lost your faith''), which is a colloquial idiom. 
    \item \textbf{Implicit Sarcasm:} ex. \<يا خاين انت عايز تشكك في حب الشعب للريس> 
    % \<>
    \includegraphics[height=1em]{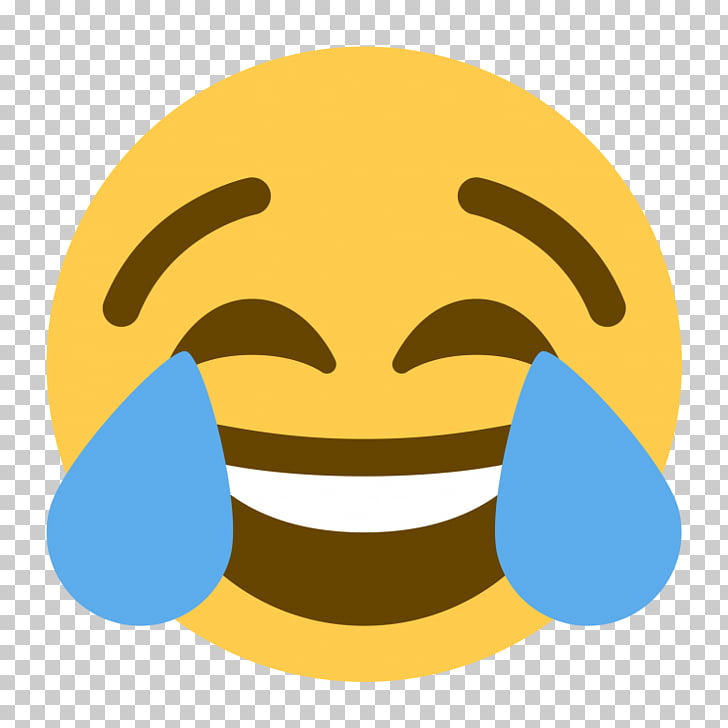} \includegraphics[height=1em]{figures/lol_emoji.jpg} (``yA xAyn Ant EAwz t\$kk fy Hb Al\$Eb llrys'' -- ``O traitor, (you) want to question people's love for the president \includegraphics[height=1em]{figures/lol_emoji.jpg}\includegraphics[height=1em]{figures/lol_emoji.jpg}'') where the author is mocking the president's popularity.
    % \squeezeupless
\end{itemize}

\paragraph{Two false negative types:} % had two types: %fell into one of the three categories:
\begin{itemize}[leftmargin=*]
    \item \textbf{Mixture of offensiveness and admiration:} ex. calling a girl a puppy \<يا كلبوبة> 
    (``yA klbwbp'' -- ``O puppy'') in a flirtatious manner. % is offensive, despite the seemingly positive connotation. %s with the seemingly complementary statement.
    \item \textbf{Implicit offensiveness:} ex. calling for cure while implying sanity:   \<وتشفي حكام بلدك من المرض> 
    (``wt\$fy HkAm bldk mn AlmrD'' -- ``and cure rulers of your country from illness'').
\end{itemize}
\section{Conclusion and Future Work}
In this paper we presented a systematic method for building an Arabic offensive language tweet dataset that does not favor specific dialects, topics, or genres.  We developed detailed guidelines for tagging the tweets as clean or offensive, including special tags for vulgar tweets and hate speech.  We tagged 10,000 tweets, which we plan to release publicly and would constitute the largest available Arabic offensive language dataset.  We characterized the offensive tweets in the dataset to determine the topics that illicit such language, the dialects that are most often used, the common modes of offensiveness, and the gender distribution of their authors.  We performed this breakdown for offensive tweets in general and for vulgar and hate speech tweets separately.  We believe that this is the first detailed analysis of its kind.  Lastly, we conducted a large battery of experiments on the dataset, using cross-validation, to establish a strong system for Arabic offensive language detection.  We showed that using an Arabic specific BERT model (AraBERT) and static embeddings trained on tweets produced competitive results on the dataset.

For future work, we plan to pursue several directions.  First, we want explore target specific offensive language, where attacks against an entity or a group may employ certain expressions that are only offensive within the context of that target and completely innocuous otherwise.  Second, we plan to examine the effectiveness of cross dialectal and cross lingual learning of offensive language.

\bibliography{anthology,eacl2021}
\bibliographystyle{acl_natbib}

\end{document}